%% file: main.tex
\pdfoutput=1
\documentclass{article}
\PassOptionsToPackage{numbers,compress}{natbib}
\usepackage[dblblindworkshop, final]{neurips_2026}
\makeatletter
\renewcommand{\@noticestring}{Accepted at the 6th Workshop on Mathematical Reasoning and AI (MATH-AI), NeurIPS 2026.}
\makeatother
\usepackage[utf8]{inputenc}
\usepackage[T1]{fontenc}
\usepackage{hyperref,url,booktabs,microtype,xcolor}
\usepackage{amsmath,amssymb,amsthm,aliascnt}
\usepackage[nameinlink,noabbrev]{cleveref}
\usepackage{enumitem,array,tabularx,tikz}
\usetikzlibrary{decorations.pathreplacing}

\newaliascnt{lemma}{theorem}

\theoremstyle{definition}

\theoremstyle{remark}

\aliascntresetthe{lemma}
\crefname{lemma}{lemma}{lemmas}
\Crefname{lemma}{Lemma}{Lemmas}
\input{experiment-data}

\title{Bolzano: From Expert-Guided Proof Search\\to Automated Open-Problem Solving}
\author{%
Adrián Zámečník\textsuperscript{1}
\And
Matěj Kripner\textsuperscript{2}
\And
Martin Koutecký\textsuperscript{1}
\And
Martin Balko\textsuperscript{3}
\AND
Jan Grebík\textsuperscript{1}
\And
Pavel Hubáček\textsuperscript{1,4}
\And
Robert Šámal\textsuperscript{1}
\And
Václav Rozhoň\textsuperscript{1}
\AND
\normalfont
\textsuperscript{1}Computer Science Institute, Charles University\\
\textsuperscript{2}Institute of Formal and Applied Linguistics, Charles University\\
\textsuperscript{3}Department of Applied Mathematics, Charles University\\
\textsuperscript{4}Institute of Mathematics, Czech Academy of Sciences
}
\begin{document}
\maketitle

\begin{abstract}
Large language models are increasingly contributing to mathematical research, where progress often depends on efficient proof search, incremental improvements and careful verification.
We describe Bolzano, a multi-agent open-source system that uses parallel prover agents with a verifier agent and maintains a human-readable research state. Initial manual use on expert-selected problems yielded 8 results whose proofs were checked by domain experts.
Motivated by these case studies, we ran Bolzano without problem-specific human guidance on about \TotalAttempts{} open problems extracted from four sets of papers, solving about 200 open problems. One experiment used papers accepted to STOC 2026, a top conference in theoretical computer science. There, we answered four questions raised in the papers, as confirmed by their authors.
\end{abstract}

\section{Introduction}
\label{sec:introduction}

Large language models (LLMs) now contribute new mathematics beyond competition benchmarks.
Early GPT-5 and Gemini case studies documented useful proof ideas, counterexamples, and extensions of existing arguments~\cite{openai2025gpt5,google2025gemini}.
More recent milestones include a disproof of the Erd\H{o}s unit-distance conjecture, checked and refined by mathematicians~\cite{alon2026unitdistance}; a collection of advances across mathematics and theoretical computer science~\cite{openai2026tenadvances};
and a reported lower bound exceeding two thirds for the proportion of zeta zeros that are simple and on the critical line~\cite{claude2026zetazeros}.
These developments motivate systems for sustained proof search and mathematical research.

Bolzano~\cite{bolzano2026,bolzano} organizes this process through parallel informal proof exploration, critique and shared mathematical notes.
Its initial study investigated eight problems whose resulting proofs were checked by domain experts.
This experience suggested a further question: can the same research loop produce useful mathematics across many open problems without individual human steering?
We study this question through four automated experiments with \TotalAttempts{} attempts.
We describe the system and how we screened and reviewed its outputs.
Across the four experiments, Bolzano found a few hundred solutions, which we sent to authors of the source papers, some of whom plan to include these results in revised versions of their papers.
While many source papers were recent arXiv preprints, four confirmed solutions came from papers accepted to STOC 2026 -- the prime conference for theoretical computer scientists.

\paragraph{Contribution and scope.}
We investigate how automated problem extraction and proof search can turn open questions in research papers into new mathematical results. Each stage of the workflow is described, from problem extraction to human review, and we report results across combinatorics and theoretical computer science.

\paragraph{Related research harnesses.}
Aletheia~\cite{aletheia2026} implements an iterative framework for natural-language proof generation, verification and revision, whereas AlphaEvolve~\cite{novikov2025ai} combines evolutionary program search with executable evaluation.
Feng et al.~\cite{feng2026erdos} study automated attempts on open Erd\H{o}s problems, with human evaluation of correctness and novelty.
Bolzano adopts a related iterative architecture, combining parallel informal proof generation with verifier feedback and a persistent shared knowledge base.
This architecture supports both expert-guided and autonomous investigations, whose mathematical outputs require subsequent human assessment.

\section{Bolzano's research loop}
\label{sec:bolzano}

Bolzano is an open-source orchestration layer around LLMs.
An \emph{agent} is a model call with a role-specific prompt and supplied research context.
The standard pipeline runs sequential \emph{research rounds}, each consisting of parallel provers, a verifier and a summarizer.
The architecture below describes this standard pipeline.

\paragraph{Explicit research state.}
For a problem task, the state entering a round contains three Markdown documents built in previous rounds:
\texttt{notes.md} stores various insights, failed approaches, conjectures, simplified proofs, etc.;
\texttt{proofs.md} contains detailed candidate proofs;
and \texttt{output.md} explains the current outcome for the researcher.
The state also includes earlier round summaries.
These files give each round a short record of earlier work.
They let agents build on earlier ideas without reading every previous response.
This makes it possible to improve a proof over several rounds instead of starting again each time.

\paragraph{Agent setup.} In all reported experiments, the prover, verifier, and summarizer operated without web search, code execution, direct filesystem access, or any tools. The research state documents were included in the agents' prompts in all runs. The reason for these decisions was to evaluate LLMs' pure reasoning capabilities over supplied mathematical text.

\paragraph{Parallel exploration.}
Each of the provers receives the task, the same current documents and optionally instructions specific to its role or round.
Provers are asked to explore proof strategies, construct examples or counterexamples, prove special cases and identify gaps.
Their model and reasoning settings can be configured individually.
They run concurrently and do not see one another's current-round responses. Responses are then collected for the verifier.
Parallel calls explore a common starting point but need not yield independent approaches or errors.

\paragraph{Critique and state update.}
The verifier receives the shared context and all prover responses for the current round.
It is prompted to assess the arguments, identify unjustified steps, reconcile compatible ideas, and decide what to retain.
Its structured response contains feedback, blocking issues and proposed updates to the three documents.
The application validates this response's schema and applies append or replace operations to the documents.
Within the automated loop, only this verifier stage updates the shared documents; provers propose material but do not overwrite the common state.

\paragraph{Summaries and continuation.}
A separate summarizer receives the verifier's response, the updated documents and previous summaries.
It produces a concise account of the round's progress and remaining questions, which is stored for display and later rounds.
It does not write to the three research documents.
After summarization, the next round is scheduled if the preset round budget remains.
A run marked completed has finished its allocated rounds; this status does not mean the problem has been solved.
In interactive use, a researcher can add instructions between rounds. In the automated experiments, no problem-specific human guidance was supplied during solving.

\paragraph{Scope of verification.}
The verifier is another LLM, not a formal proof checker.
An accepted but incorrect lemma may propagate through persistent memory, and later agents may repeat it with increasing confidence.
Bolzano's documents therefore remain candidate mathematics until checked by a human or a formal system.

\section{Initial use with domain experts}
\label{sec:problems}

The initial Bolzano report~\cite{bolzano2026} documents eight problems solved by the system and checked by domain experts, with strategic guidance supplied in some cases.
It classifies six results as publishable research and five as essentially autonomous under the taxonomy of Feng et al.~\cite{aletheia2026}; autonomy here concerns argument generation after human problem preparation.
The full problem statements, results, and proofs appear in the original report~\cite{bolzano2026}.

These resolved problems motivated a scalable workflow: allocate a fixed research budget to each problem, retain counterexamples and partial advances as well as complete proofs and concentrate human attention on the most promising outputs.

\section{Four automated experiments}
\label{sec:experiments}

We applied this workflow to four collections of papers: newly released arXiv papers in \texttt{math.CO} and \texttt{cs.DS}, STOC 2026 accepted papers, earlier FOCS, SODA, and STOC papers and work associated with the Midsummer Combinatorial Workshop 2026 and earlier.
A separate LLM-based workflow extracted open problems from these papers with the definitions and statements needed to make them self-contained.
Each extracted problem initiated one investigation, as shown in \Cref{tab:experiments}.
Problems may overlap between collections.

\begin{table}[htbp]
\caption{Automated experiments. Different collections use different run configurations. All problem runs used only one prover set to maximum reasoning effort available for the model.}
\label{tab:experiments}
\centering
\small
\setlength{\tabcolsep}{8pt}
\begin{tabularx}{\linewidth}{@{}Xrcccc@{}}
\toprule
Collection & Problems & Model & Rounds & Promising\footnotemark[1] & Solved\footnotemark[2] \\
\midrule
arXiv: \texttt{math.CO}, \texttt{cs.DS} & \ArxivAttempts & \ArxivModel & \ArxivRounds & \ArxivReviewed & \ArxivSolved \\
STOC 2026 & \StocAttempts & \StocModel & \StocRounds & \StocReviewed & \StocSolved \\
Earlier FOCS/SODA/STOC & \EarlierAttempts & \EarlierModel & \EarlierRounds & \EarlierReviewed & \EarlierSolved \\
Midsummer Combinatorial Workshop & \McwAttempts & \McwModel & \McwRounds & \McwReviewed & \McwSolved \\
\midrule
Total & \TotalAttempts & -- & -- & \TotalReviewed & \TotalSolved \\
\bottomrule
\end{tabularx}
\end{table}
\footnotetext[1]{Outputs flagged by Bolzano as candidate resolutions and selected for
human assessment of correctness and relevance to the source question. Outputs not flagged as promising were not systematically reviewed.}
\footnotetext[2]{Except for STOC 2026, the solved counts are estimates. These estimates are partially informed by LLM-based assessments of whether the output addresses the source question and whether its argument is substantially correct. We received feedback from authors on many of the results, which suggests that these estimates are reasonable. For STOC 2026, we report only the four results that we checked carefully and that authors of the source papers verified.}

\paragraph{Open problem extraction.}
We initially used LLMs to identify explicit open problems, questions and conjectures in the collected papers. This process evolved into a more structured workflow using a general-purpose agent harness with access to local paper files and web search. The agents read the papers and look for explicitly stated problems. They then check whether the paper itself resolves the problem and perform a literature search to see if the problem is still open. Candidates identified as already resolved were excluded. For each retained problem, the agents assembled a self-contained task with necessary definitions, assumptions, and related work. These tasks were then supplied to Bolzano.  

\paragraph{Run configuration and artifacts.}
For a collection, the run configuration specifies the prover, verifier, and summarizer models, their reasoning settings, and number of research rounds.
No problem-specific human instructions are added during solving.
The result records the extracted task, final output, detailed proof record, and, when available, research notes and the source paper.
These artifacts allow postprocessing to examine the mathematical argument separately from the run's completion status.

\paragraph{Screening and source checks.}
LLM postprocessing reads the output and relevant proof material to identify substantive claims: proofs, counterexamples, algorithms or improved bounds.
The run's own status is not evidence of progress and a persistent registry identifies duplicate packages.
Promising claims are compared with the actual source question, including its hypotheses, quantifiers, and conventions, and assessed for proof gaps, restricted scope and overlap with known results.
This check is necessary because a correct proof of an inaccurately extracted statement may not advance the original problem.

\paragraph{Structured audit and human review.}
Selected candidates were rewritten as mathematical notes; some received a further LLM audit using the note and source text, with the extracted problem treated only as a non-authoritative hint.
The audit requires two conditions: the claimed problem is supported by the source under its intended reading and the proof is substantially correct.
The source assessment requests a precise question or conjecture and supporting text.
A failed or unresolved condition produces a negative decision.
The audit also requests structured assessments of resolution completeness, scope limitations, issue severity, fixability, and review priority.
Thus, an audit pass can describe substantial partial progress.
These records guide closer human inspection and, where available, assessment by the original problem authors.
Our expertise does not cover all relevant fields. Screening and audit protocols also evolved across the experiments.

\subsection{Four results from STOC 2026}

We selected four results from the STOC 2026 experiment and sent them to authors of the source papers.
The authors confirmed the results in private correspondence in June and July 2026.\footnote[3]{For the results on convex quartics~\cite{slot2025hesse} and biased CAT states~\cite{parham2025magic}, the authors reviewed only the main idea, not the full proof.}

\paragraph{Planted bicliques.}
Bolzano found a one-pass algorithm that uses $O(\!\log n)$ bits of memory to detect a semi-random planted biclique when both sides have size at least $Cn^{2/3}$~\cite{garg2026memory}.
This matches the known lower bound up to logarithmic factors.

\paragraph{Monotone rank programs.}
Bolzano found an exponential separation between monotone rank programs and monotone span programs~\cite[Section~1.3]{cavalar2025negations}.
For $F_m(a,b)=\bigvee_{i=1}^{m}(a_i\wedge b_i)$, there is a monotone span program with $2m$ rows over every field, while every column-full-rank monotone rank program needs exactly $m2^m$ rows.

\paragraph{Convex quartics.}
Bolzano constructed a strongly convex quartic polynomial in two variables with rational coefficients whose zero sublevel set consists of one irrational point~\cite{slot2025hesse}.
This gives a convex quartic feasibility problem that has a solution but no rational witness.

\paragraph{Biased CAT states.}
Bolzano proved an $\Omega(\log n)$ circuit-depth lower bound for both joining two biased CAT states and splitting one into two, including the entropy-balanced case $H(\alpha)=2H(\beta)$ left open in the source paper~\cite{parham2025magic}.

\section{Discussion}
\label{sec:conclusion}

Bolzano connects parallel proof exploration to a persistent research record that supports both human steering and unattended runs.
The manually checked case studies establish that this workflow can contribute to research and the four experiments extend its reach beyond individually prepared questions.
They also expose the remaining bottleneck: assessing the correctness, intended scope, and novelty of candidate results.
Observed failures include missing assumptions, proofs addressing weaker versions of the original questions, invalid arguments and rediscoveries of known results.
Screening can itself miss useful output, so unflagged runs cannot all be declared unsolved.
The experiments use different problem collections, models, budgets and include no comparison with direct use of the underlying models.
They demonstrate a practical research workflow, while leaving its comparative efficiency and the final yield of verified solutions to further evaluation.

\label{main-text-end}
\input{acknowledgments}
\bibliographystyle{plainnat}
\bibliography{ref}
\end{document}

%% file: experiment-data.tex
\newcommand{\ArxivAttempts}{1{,}600}
\newcommand{\StocAttempts}{420}
\newcommand{\EarlierAttempts}{900}
\newcommand{\McwAttempts}{880}
\newcommand{\TotalAttempts}{3{,}800}

\newcommand{\ArxivModel}{GPT-5.5}
\newcommand{\StocModel}{GPT-5.5}
\newcommand{\EarlierModel}{GPT-5.6 Sol}
\newcommand{\McwModel}{GPT-5.6 Sol}

\newcommand{\ArxivRounds}{4}
\newcommand{\StocRounds}{4}
\newcommand{\EarlierRounds}{2}
\newcommand{\McwRounds}{2}

\newcommand{\ArxivReviewed}{250}
\newcommand{\StocReviewed}{41}
\newcommand{\EarlierReviewed}{57}
\newcommand{\McwReviewed}{137}
\newcommand{\TotalReviewed}{485}

\newcommand{\ArxivSolved}{90}
\newcommand{\StocSolved}{4}
\newcommand{\EarlierSolved}{40}
\newcommand{\McwSolved}{80}
\newcommand{\TotalSolved}{214}

%% file: acknowledgments.tex
\begin{ack}
Kripner was supported by Charles University, project GA UK no. 458326. Koutecký was partially supported by Charles University project UNCE 24/SCI/008, by the ERC-CZ project LL2406 of the Ministry of Education of the Czech Republic, and by the project 25-17221S of the Czech Science Foundation (GACR). Grebík, Rozhoň, and Zámečník were supported by GACR,
JUNIOR STAR project no. 26-23599M.
Šámal was supported by grant 25-16627S of GACR.
\end{ack}

%% file: main.bbl
\begin{thebibliography}{14}
\providecommand{\natexlab}[1]{#1}
\providecommand{\url}[1]{\texttt{#1}}
\expandafter\ifx\csname urlstyle\endcsname\relax
  \providecommand{\doi}[1]{doi: #1}\else
  \providecommand{\doi}{doi: \begingroup \urlstyle{rm}\Url}\fi

\bibitem[Alon et~al.(2026)Alon, Bloom, Gowers, Litt, Sawin, Shankar, Tsimerman,
  Wang, and Wood]{alon2026unitdistance}
Noga Alon, Thomas~F. Bloom, W.~T. Gowers, Daniel Litt, Will Sawin, Arul
  Shankar, Jacob Tsimerman, Victor Wang, and Melanie~Matchett Wood.
\newblock Remarks on the disproof of the unit distance conjecture, 2026.
\newblock URL
  \url{https://cdn.openai.com/pdf/74c24085-19b0-4534-9c90-465b8e29ad73/unit-distance-remarks.pdf}.

\bibitem[Balko et~al.(2026)Balko, Greb{\'i}k, Hub{\'a}{\v{c}}ek, Kouteck{\'y},
  Kripner, Rozho{\v{n}}, {\v{S}}{\'a}mal, and
  Z{\'a}me{\v{c}}n{\'i}k]{bolzano2026}
Martin Balko, Jan Greb{\'i}k, Pavel Hub{\'a}{\v{c}}ek, Martin Kouteck{\'y},
  Mat{\v{e}}j Kripner, V{\'a}clav Rozho{\v{n}}, Robert {\v{S}}{\'a}mal, and
  Adri{\'a}n Z{\'a}me{\v{c}}n{\'i}k.
\newblock {Bolzano}: Case studies in {LLM}-assisted mathematical research,
  2026.
\newblock \href{https://arxiv.org/abs/2604.16989}{arXiv:2604.16989}.

\bibitem[{Bolzano Team}(2025)]{bolzano}
{Bolzano Team}.
\newblock Bolzano, 2025.
\newblock URL \url{https://bolzano.app/}.

\bibitem[Bubeck et~al.(2025)Bubeck, Coester, Eldan, Gowers, Lee, Lupsasca,
  Sawhney, Scherrer, Sellke, Spears, Unutmaz, Weil, Yin, and
  Zhivotovskiy]{openai2025gpt5}
S\'{e}bastien Bubeck, Christian Coester, Ronen Eldan, Timothy Gowers, Yin~Tat
  Lee, Alexandru Lupsasca, Mehtaab Sawhney, Robert Scherrer, Mark Sellke,
  Brian~K. Spears, Derya Unutmaz, Kevin Weil, Steven Yin, and Nikita
  Zhivotovskiy.
\newblock Early science acceleration experiments with {GPT}-5, 2025.
\newblock \href{https://arxiv.org/abs/2511.16072}{arXiv:2511.16072}.

\bibitem[Cavalar et~al.(2025)Cavalar, Fabris, Mukhopadhyay, Srinivasan, and
  Yehudayoff]{cavalar2025negations}
Bruno Cavalar, Th{\'e}o~Bor{\'e}m Fabris, Partha Mukhopadhyay, Srikanth
  Srinivasan, and Amir Yehudayoff.
\newblock Negations are powerful even in small depth, 2025.
\newblock \href{https://arxiv.org/abs/2512.19515}{arXiv:2512.19515}.

\bibitem[{Claude}(2026)]{claude2026zetazeros}
{Claude}.
\newblock More than two thirds of the zeros of the {Riemann} zeta function are
  simple and on the critical line.
\newblock Anthropic technical report, 2026.
\newblock URL
  \url{https://www-cdn.anthropic.com/95c246936988e43127bc6b2ceb7077c1dad2d68e.pdf}.

\bibitem[Feng et~al.(2026{\natexlab{a}})]{aletheia2026}
Tony Feng et~al.
\newblock Towards autonomous mathematics research, 2026{\natexlab{a}}.
\newblock \href{https://arxiv.org/abs/2602.10177}{arXiv:2602.10177}.

\bibitem[Feng et~al.(2026{\natexlab{b}})]{feng2026erdos}
Tony Feng et~al.
\newblock Semi-autonomous mathematics discovery with {G}emini: A case study on
  the {Erd\H{o}s} problems, 2026{\natexlab{b}}.
\newblock \href{https://arxiv.org/abs/2601.22401}{arXiv:2601.22401}.

\bibitem[Garg et~al.(2026)Garg, Hastings, Pabbaraju, and
  Sharan]{garg2026memory}
Sumegha Garg, Jabari Hastings, Chirag Pabbaraju, and Vatsal Sharan.
\newblock A unified approach to memory-sample tradeoffs for detecting planted
  structures, 2026.
\newblock \href{https://arxiv.org/abs/2603.00770}{arXiv:2603.00770}.

\bibitem[Novikov et~al.(2025)]{novikov2025ai}
Alexander Novikov et~al.
\newblock {AlphaEvolve}: A coding agent for scientific and algorithmic
  discovery, 2025.
\newblock \href{https://arxiv.org/abs/2506.13131}{arXiv:2506.13131}.

\bibitem[{OpenAI}(2026)]{openai2026tenadvances}
{OpenAI}.
\newblock Ten advances in mathematics and theoretical computer science, 2026.
\newblock URL \url{https://cdn.openai.com/pdf/ten-proofs-oai.pdf}.

\bibitem[Parham(2025)]{parham2025magic}
Natalie Parham.
\newblock Quantum circuit lower bounds in the magic hierarchy, 2025.
\newblock \href{https://arxiv.org/abs/2504.19966}{arXiv:2504.19966}.

\bibitem[Slot et~al.(2025)Slot, Steurer, and Wiedmer]{slot2025hesse}
Lucas Slot, David Steurer, and Manuel Wiedmer.
\newblock Hesse's redemption: Efficient convex polynomial programming, 2025.
\newblock \href{https://arxiv.org/abs/2511.03440}{arXiv:2511.03440}.

\bibitem[Woodruff et~al.(2026)Woodruff, Cohen-Addad, Jain, Mao, Zuo,
  et~al.]{google2025gemini}
David~P. Woodruff, Vincent Cohen-Addad, Lalit Jain, Jieming Mao, Song Zuo,
  et~al.
\newblock Accelerating scientific research with {G}emini: Case studies and
  common techniques, 2026.
\newblock \href{https://arxiv.org/abs/2602.03837}{arXiv:2602.03837}.

\end{thebibliography}
